\documentclass[10pt,twocolumn,letterpaper]{article}

\usepackage{iccv}
\usepackage{epsfig}
\usepackage{graphicx}
\usepackage{amsmath}
\usepackage{amssymb}
\usepackage{multirow, eucal}
\usepackage{color}
\usepackage{float}

\usepackage[margin=18pt,font=small,labelfont=bf,labelsep=endash,tableposition=top]{caption}

\iccvfinalcopy

\begin{document}

\title{Towards Context-aware Interaction Recognition\thanks{The first two authors contributed equally to this work.
    This work was in part supported by an ARC Future Fellowship to C. Shen.
  Corresponding author: C. Shen (e-mail: chunhua.shen@adelaide.edu.au).
}
}

\author{ { Bohan Zhuang,   Lingqiao Liu,   Chunhua Shen,   Ian Reid}\\
School of Computer Science, University of Adelaide, Australia}

\maketitle
\begin{abstract}
	Recognizing how objects interact with each other is a crucial task in visual recognition.
  If we define the context of the interaction to be the objects involved, then most current methods can be categorized as either:
  (i) training a single classifier on the combination of the interaction and its context; or
  (ii) aiming to recognize the interaction independently of its explicit context. Both methods suffer limitations: the former scales poorly with the number of combinations and fails to generalize to unseen combinations, while the latter often leads to poor interaction recognition performance due to the difficulty of designing a context-independent interaction classifier. To mitigate those drawbacks, this paper proposes an alternative, context-aware interaction recognition framework. The key to our method is to explicitly construct an interaction classifier which combines the context, and the interaction.  The context is encoded via word2vec into a semantic space, and is used to derive a classification result for the interaction.

  The proposed method still builds one classifier for one interaction (as per type (ii) above), but the classifier built is adaptive to context via weights which are context dependent. The benefit of using the semantic space is that it naturally leads to zero-shot generalizations in which semantically similar contexts (subject-object pairs) can be recognized as suitable contexts for an interaction, even if they were not observed in the training set. Our method also scales with the number of interaction-context pairs since our model parameters do not increase with the number of interactions. Thus our method avoids the limitation of both approaches. We demonstrate experimentally that the proposed framework leads to improved performance for all investigated interaction representations and datasets.

\end{abstract}

\tableofcontents
\newpage

\begin{figure*}[tbp]
	\centering
	\resizebox{1.0\linewidth}{!}
	{\begin{tabular}{c}
			\includegraphics{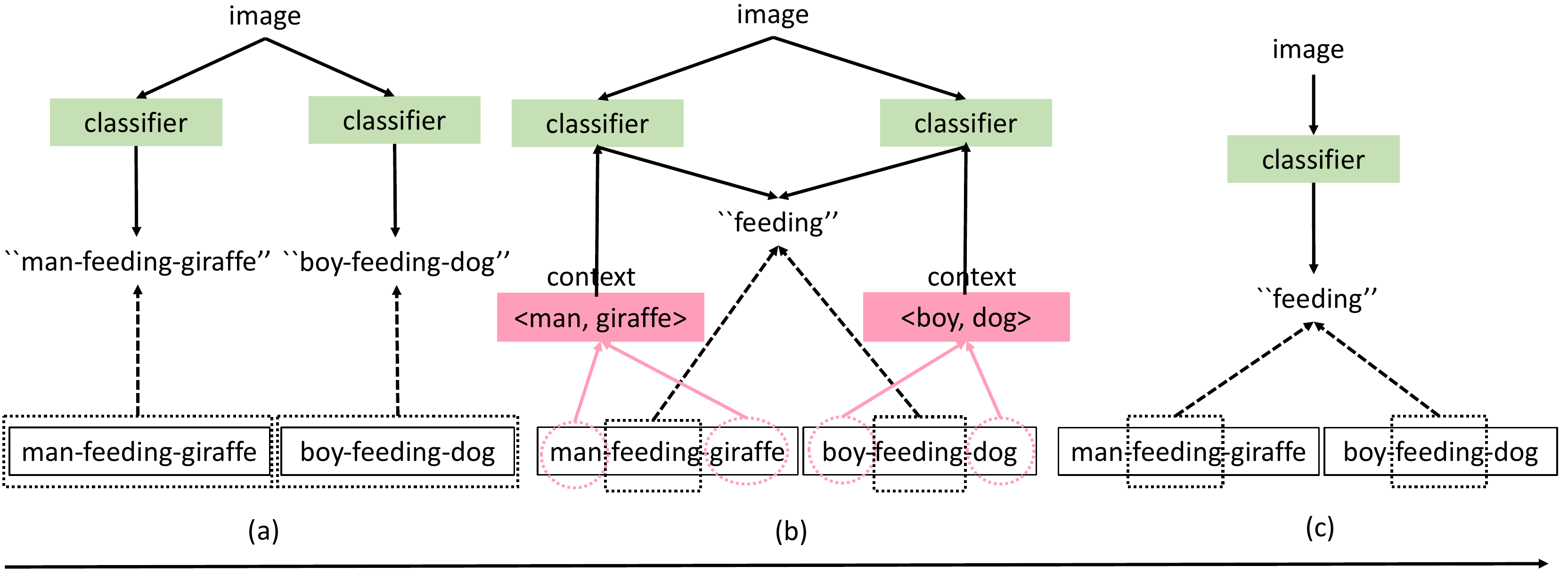}
		\end{tabular}
	}
	\caption{Comparison of two baseline interaction recognition methods and the proposed approach. The two baseline methods take two extremes. For one extreme, (a) treats the combination of the interaction and its context as a single class. For another extreme, (c) classifies the interaction separately from its context. Our method (b) lies somewhere between (a) and (c). We still build one classifier for each interaction but the classifier parameter is also adaptive to the context of the interaction, as shown in the example in (b).}
	\label{fig:overview}
\end{figure*}

\section{Introduction}
Object interaction recognition is a fundamental problem in computer vision and it can serve as a critical component for solving many visual recognition problems such as action recognition \cite{mallya2016learning, ramanathan2015learning, wang2015action, bilen2016dynamic, Zhang_2016_CVPR}, visual phrase recognition \cite{hu2016modeling, rohrbach2016grounding, li2017vip}, sentence to image retrieval \cite{ma2015multimodal, karpathy2015deep} and visual question answering \cite{wu2016ask, lu2016knowing, wu2016value}. Unlike object recognition in which the object appearance and its class label have a clear association, the interaction patterns, e.g. ``eating'', ``playing'', ``stand on'',  usually have a vague connection to visual appearance. This phenomenon is largely caused by the same interaction being involved with different objects as its context, i.e. the subject and object of an interaction type. For example, ``cow eating grass'' and ``people eating bread'' can be visually dissimilar although both of them have the same interaction type ``eating''. Thus the subject and object associated with the interaction -- also known as the \textit{context} of the interaction -- could play an important role in interaction recognition.

In existing literature, there are two ways to model the interaction and its context. The first one treats the combination of interaction and its context as a single class.  For example, in this approach, two classifiers will be built to classify ``cow eating grass" and ``people eating bread." To recognize the interaction ``eating'', images that are classified as either ``cow eating grass'' or ``people eating bread'' will be considered as having interaction ``eating". This treatment has been widely used in defining action (interaction) classes in many action (interaction) recognition benchmarks~\cite{mallya2016learning, ramanathan2015learning, wang2015action, bilen2016dynamic, Zhang_2016_CVPR}. This approach, however, suffers from poor scalability and generalization ability. The number of possible combinations of the interaction and its context can be huge, and thus it is very inefficient to collect training images for each combination. Also, this method fails to generalize to an unseen combination even if both its interaction type and context are seen in the training set.

To handle these drawbacks, another way is to model the interaction and the context separately~\cite{lu2016visual, desai2011discriminative, gupta2008beyond, sadeghi2015viske}. In this case, the interaction is classified independently of its context, which can lead to poor recognition performance due to the difficulty of associating the interaction with certain visual appearance in the absence of context information. To overcome the imperfection of interaction classification, some recent works employ techniques such as language priors \cite{lu2016visual} or structural learning \cite{li2017vip, Liang2017VRD} to avoid generating an unreasonable combination of interaction and context.  However, the context-independent interaction classifier is still used as a building block, and this prevents the system from gaining more accurate recognition from visual cues.

The solution proposed in this paper aims to overcome the drawbacks of both methods. To avoid the explosion of the number of classes, we still separate the classification of the interaction and the context into two stages. However, different to the second method, the interaction classifier in our method is designed to be adaptive to its context. In other words, for the same interaction, different contexts will result in different classifiers and our method will encourage interactions with similar contexts to have similar classifiers. By doing so, we can achieve context-aware interaction classification while avoiding treating each combination of context and interaction as a single class. Based on this framework, we investigate various feature representations to characterize the interaction pattern. We show that our framework can lead to performance improvements for all the investigated feature representations. Moreover,  we augment the proposed framework with an attention mechanism, which leads to further improvements and yields our best performing recognition model.  Through extensive experiments, we demonstrate that the proposed methods achieve superior performance over competing methods.

\section{Related work}

\emph{Action recognition:} Action is one of the most important interaction patterns and action recognition in images/videos has been widely studied ~\cite{mallya2016learning, ramanathan2015learning, wang2015action, bilen2016dynamic, Zhang_2016_CVPR}. Various action recognition datasets such as Stanford 40 actions~\cite{yao2011human}, UCF-101~\cite{soomro2012ucf101} and HICO~\cite{chao2015hico} have been proposed, but most of them focus on actions (interactions) with limited number of context. For example, in the relatively large HICO~\cite{chao2015hico} dataset, there are only 600 categories of human-object interactions. Thus the interplay of the interaction and its context has not been explored in the works of this direction.

\emph{Visual relationships:} Some recent works focus on the detection of visual relationships. A visual relationship is composed of an interaction and its context, i.e. subject and object. Thus this direction is most relevant to this paper. In fact, the interaction recognition can be viewed as the most challenging part of the visual relationship detection. Some recent works in visual relationship detection have made progress in improving the detection performance and the detection scalability.  The work in~\cite{lu2016visual} leveraged language priors to produce relationship detections that make sense to human beings. The latest approaches~\cite{Liang2017VRD, li2017vip, zhang2017visual} attempt to learn the visual relationship detector in an end-to-end manner and explicitly reason the interdependency among relationship components at the visual feature level.

\emph{Language-guided visual recognition:} Our method uses language information to guide the visual recognition. This corresponds to the recent trend in utilizing language information for benefiting visual recognition. For example, language information has also been incorporated in phrase grounding~\cite{plummer2016phrase, hu2016modeling, rohrbach2016grounding} tasks. In~\cite{hu2016modeling, rohrbach2016grounding}, attention model is employed to extract linguistic cues from phrases. Language guided attention has also been widely used in visual question answering~\cite{donahue2015long,karpathy2015deep, malinowski2015ask, ren2015image} and has recently been applied to one-shot learning~\cite{vinyals2016matching}.

\section{Methods}

\subsection{Context-aware interaction classification framework}

In general, an interaction and its context can be expressed as a triplet $\left\langle \emph{O1-P-O2} \right\rangle$, where $P$ denotes the interaction, and $O1$ and $O2$ denote its subject and object respectively. In our study, we assume the interaction context (\emph{O1,O2}) has been detected by a detector (i.e. we are given bounding boxes and lables for both subject $O1$ and object $O2$) and the task we are addressing is to classify their interaction type $P$. To recognize the interaction, existing works take two extremes in designing the classifier. One is to directly build a classifier for each $P$ and assume that the same classifier applies to $P$ with different context. Another takes the combination of $\left\langle \emph{O1-P-O2} \right\rangle$ as a single class and build a classifier for each combination. As discussed in the introduction section, the former does not fully leverage the contextual information for interaction recognition while the latter suffers from the scalability and generalization issues. Our proposed method lies between those two extremes. Specifically, we still allocate one classifier for each interaction type, however we make the classifier parameters adaptive to the context of the interaction. In other words, the classifier is a function of the context. The schematic illustration of this idea is shown in Figure \ref{fig:overview}.

Formally, we assume that the interaction classifier takes a linear classifier form $y_p = \mathbf{w}_p^{\top}\phi(I),~~\mathbf{w}_p \in \mathbb{R}^d$, where $y_p$ is the classification score for the $p$-th interaction and $\phi(I)$ is the feature representation extracted from the input image. The classifier parameters for the $p$-th interaction $\mathbf{w}_p$ are a function of $(O1,O2)$, that is, the context of the $p$-th interaction. It is designed as the summation of the following two terms:
\begin{equation}  \label{Eq:combine}
	\mathbf{w}_p(O1,O2) = \mathbf{\bar{w}}_p + r_p(O1,O2),
\end{equation}
where the first term $\mathbf{\bar{w}}_p$ is independent of the context; it plays a role which is similar to the traditional context-independent interaction classifier. The second term $r_p(O1,O2)$ can be viewed as an auxiliary classifier generated from the information of context $(O1,O2)$. Note that the summation of two classifiers has been widely used in transfer learning \cite{patricia2014learning, arnold2007comparative, do2005transfer} and multi-task learning \cite{evgeniou2004regularized, parameswaran2010large}, e.g. one term corresponds to the classifier learned in the target domain and another corresponds to the classifier learned in the source domain.

\begin{figure*}[t]
	\centering
	\resizebox{0.740\linewidth}{!}
	{\includegraphics{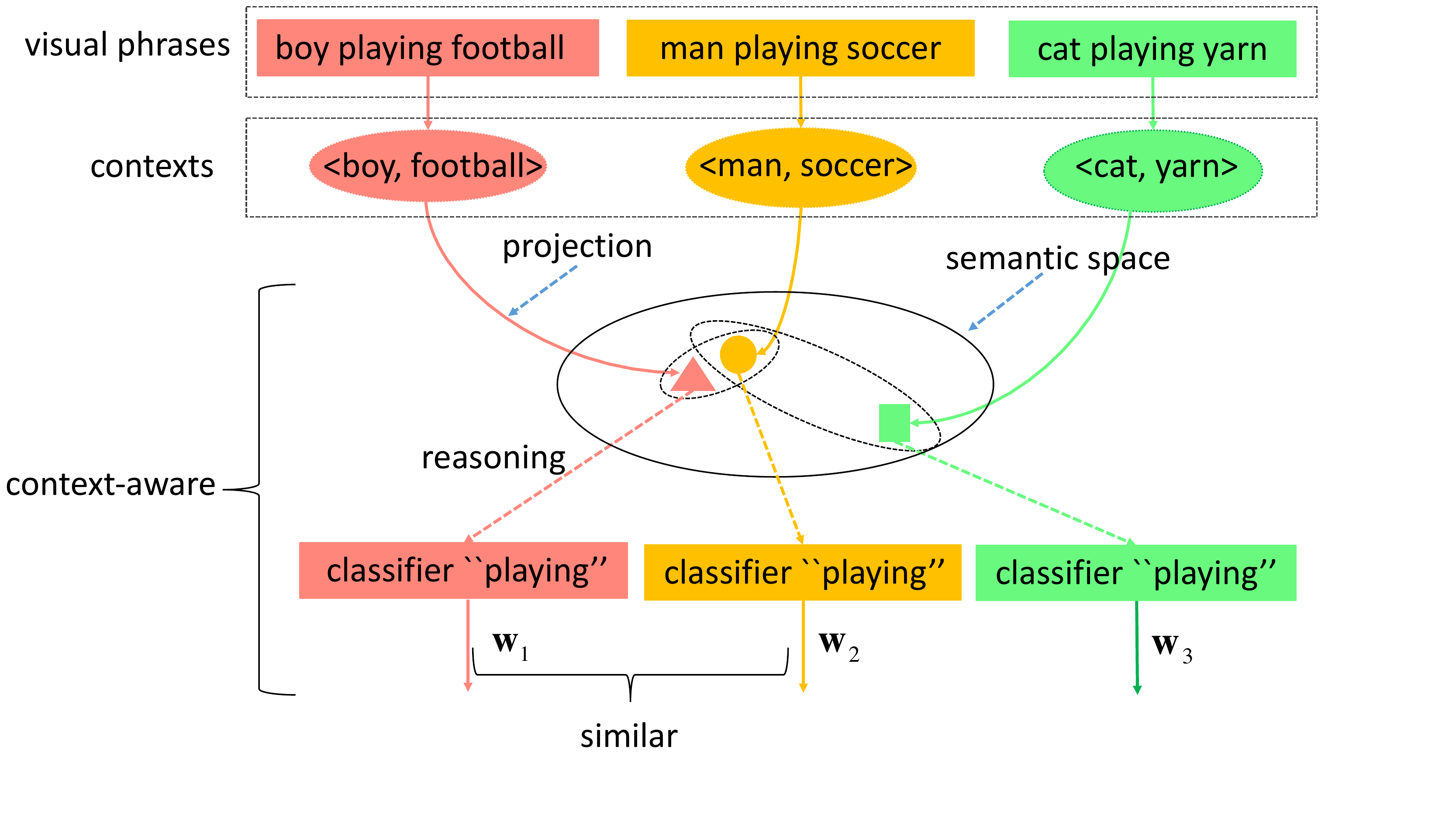}}
	\caption{An example of the proposed context-aware model. The same interaction ``playing'' is associated with various contexts. The contexts of the first two phrases are semantically similar, resulting in two similar context-aware classifiers. Since the last two contexts are far away from each other in the semantic space, their corresponding context-aware classifiers may not similar despite sharing the same label. In this way, we explicitly consider the visual appearance variations introduced by changing context, thus more accurate and generalizable interaction classifiers can be learned.}
	\label{fig:context}
\end{figure*}

Intuitively, for two interaction-context combinations, if both of them share the same interaction and their contexts are similar, the interaction in those combinations tends to be associated with similar visual appearance. For example, $\left\langle \emph{boy, playing, football} \right\rangle$ and $\left\langle \emph{man, playing, soccer} \right\rangle$ share similar context, so the interaction ``playing'' should suggest similar visual appearance for these two combinations. This inspires us to design $\mathbf{w}_p(O1,O2)$ to allow semantically similar contexts to generate similar interaction classifiers, as demonstrated in Figure \ref{fig:context}. To realize this idea, we first represent the object and subject through their word2vec embedding which maps semantically similar words into similar vectors and then generate the auxiliary classifier $r_p$ by concatenating their embeddings. Formally, $r_p$ is designed as:
\begin{align} \label{eq:relation_vec}
	r_p(O1,O2) = {{\mathbf{V}_p}}f(\mathbf{Q}E(O1,\,O2)),
\end{align}
where $E(O1,\,O2) \in \mathbb{R}^{2e}$ is the concatenation of the $e$-dimensional word2vec embeddings of $(O1,O2)$, and $\mathbf{Q} \in \mathbb{R}^{m \times 2e}$ is a projection matrix to project $E(O1,\,O2)$ to a low-dimensional (e.g. 20) semantic embedding space. $f(\cdot)$ is the RELU function and $\bf{V}_p$ transforms the context embedding to the auxiliary classifier. Note that  $\mathbf{V}_p$ and $\mathbf{\bar{w}}_p$ in Eq.~(\ref{Eq:combine}) are distinct per interaction type $p$ while the projection matrix $\mathbf{Q}$ is shared across all interactions.  All of these parameters are learnt at training time.

\noindent\textbf{Remark:} Many recent works \cite{Liang2017VRD, li2017vip, zhang2017visual, plummer2016phrase} on visual relationship detection takes a structural learning alike formulation to simultaneously predict $O1,O2$ and $P$. The unary term used in their framework is still a context-independent classifier and such choice may lead to poor recognition accuracy in identifying interaction from the visual cues. To improve these techniques, one could replace their unary terms with our context-aware interaction recognition module. On the other hand, their simultaneous prediction framework could also benefit our method in achieving better visual relationship performance. Since our focus is to study the interaction part, we do not pursue this direction in this paper and leave it for future work.

\subsection{Feature representations for interactions recognition}
One remaining issue in implementing the framework in Eq. (\ref{Eq:combine}) is the design of $\phi(I)$, that is, the feature representation of the interaction. It is clear that the choice of the feature representation can have significant impact on the interaction prediction performance. In this section, we investigate two types of feature representations to characterize the interaction. We evaluate these feature representations in Sec.~\ref{sec:VRD}.

\subsubsection{Spatial feature representation} \label{sec:spatial}
Our method assumes that the context has been detected and therefore the interaction between the subject and the object could be characterized by the spatial features of the detection bounding boxes. These kind of features have been previously employed ~\cite{hu2016modeling, plummer2016phrase, zhang2017visual} to recognize the visual relationship of objects. In our study, we use both the spatial features from each bounding box and the spatial features from their mutual relationship. Formally, let $(x,y,w,h)$ and $(x',y',w',h')$ be the bounding box coordinates of the \emph{subject} and \emph{object}, respectively. Given the bounding boxes, the spatial feature for a single box is a 5-dimentional vector represented as $[\frac{x}{{{W_I}}},\frac{y}{{{H_I}}},\frac{{x + w}}{{{W_I}}},\frac{{y + h}}{{{H_I}}},\frac{{{S_b}}}{{{S_I}}}]$, where $S_b$ and $S_I$ are the areas of region $b$ and image $I$, $W_I$ and $H_I$ are the width and height of the image $I$. And the pairwise spatial vector is denoted as $[\frac{{x - x'}}{{w'}},\frac{{y - y'}}{{h'}},\log \frac{w}{{w'}},\log \frac{h}{{h'}}]$. We concatenate them together to get a 14-dimentional feature representation (using both subject and object bounding boxes). Then the spatial feature directly passes through the context-aware classifier defined in Eq.~(\ref{Eq:combine}) for the interaction classification.

\subsubsection{Appearance feature representation}\label{sec:appearance}
Besides spatial features, we can also use appearance features, e.g. the activations of a deep neural network to depict the interaction. In our study, we first crop the union region of the subject and object bounding boxes, and rescale the region to $224 \times 224 \times 3$ as the input of a VGG-16~\cite{simonyan2014very} CNN. We then apply the mean-pooling to the activations of the $conv5\_3$ layer as our feature representation $\phi(I)$. This feature is then fed into our context-aware interaction classifier in Eq.~(\ref{Eq:combine}). To improve the performance, we treat the  context-aware interaction classifier as a newly added layer and fine-tune this layer with the VGG-16 net in an end-to-end fashion.

\subsection{Improving appearance representation with attention and context-aware attention} \label{sec:attention}
The discriminative visual cues for interaction recognition may only appear in a small region of the input image or the image region. For example, to see if ``man riding bike'' occurs, one may need to focus on the region near human feet and bike pedal. This consideration motivates us to use attention module to encourage the network ``focus on'' discriminative regions. Specially, we can replace the mean-pooling layer in Sec.~\ref{sec:appearance} with an attention-pooling layer.

Formally, let ${{\bf{h}}_{ij}} \in {R^c}$ denote the last convolutional layer activations at the spatial location $(i,j)$, where $i=1,2,...,M$ and $j=1,2,...,N$ are the coordinates of the feature map and $M$, $N$ are the height and width of the feature map respectively, c is the number of channels. The attention pooling layer pools the convolutional layer activations into a $c$-dimensional vector through:
\begin{align}
	\begin{array}{l}
		{\bar{a}(\mathbf{h}_{ij})} = \frac{{a(\mathbf{h}_{ij}) + \varepsilon }}{{\sum\limits_i {\sum\limits_j {(a(\mathbf{h}_{ij}) + \varepsilon )} } }},\\
		{\widetilde {\mathbf{h}}} = \frac{1}{MN} \sum\limits_{ij}{\bar{a}(\mathbf{h}_{ij})}  {{\mathbf{h}}_{ij}},
	\end{array}
\end{align}
where $a(\mathbf{h}_{ij})$ is the attention generation function which produces an attention value for each location $(i,j)$. The attention value is then normalized ($\varepsilon$ is a small constant) and used as a weighting factor to pool the convolutional activations ${\mathbf{h}}_{ij}$. We consider two designs of $a(\mathbf{h}_{ij})$.

\noindent\textbf{Direct attention}:
The first attention generation function is simply designed as $a(\mathbf{h}_{ij}) = f(\mathbf{w}_{att}^{\top} \mathbf{h}_{ij} + b)$, where ${{\bf{w}}_{att}}$ and $b$ are the weight and bias of the attention model.

\begin{figure*}[h!]
	\centering
  \resizebox{0.810\linewidth}{!}
	{\includegraphics{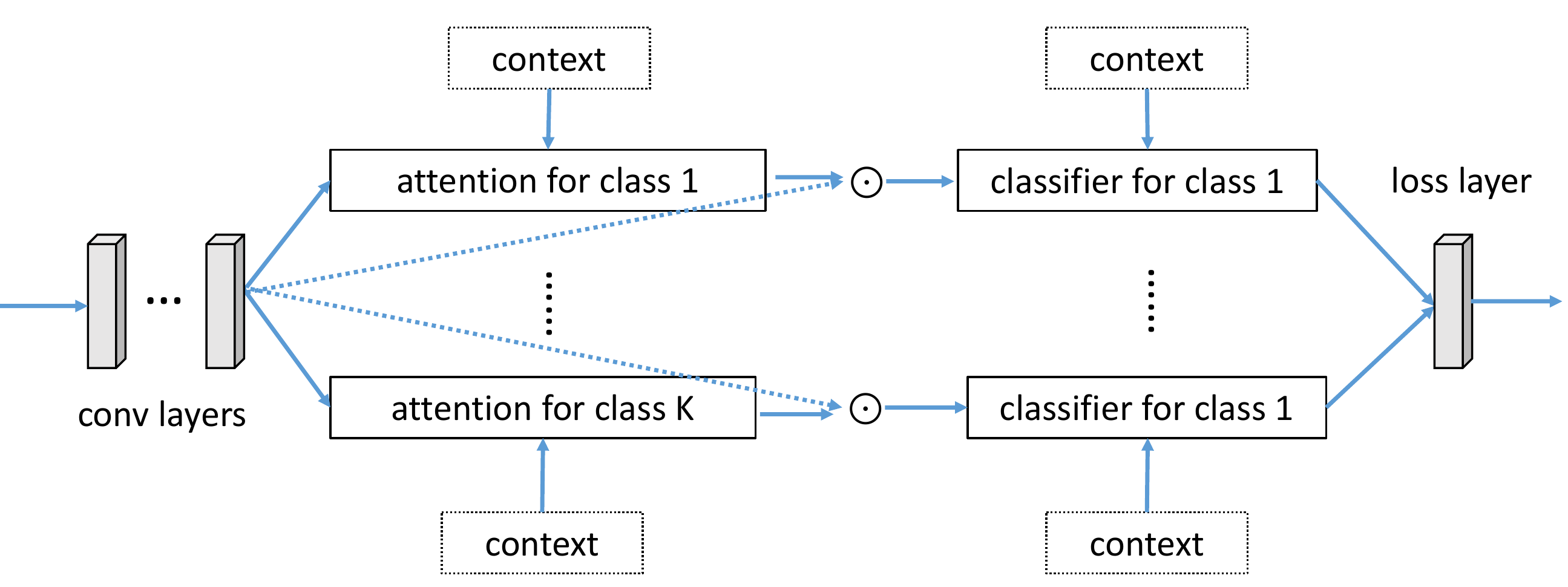}}
	\caption{Detailed illustration of the context-aware attention model. For each interaction class, there is a corresponding attention model imposed on the feature map to select the interaction-specific discriminative feature regions. Different attention-pooling vectors will be generated for different interaction classes. The generated pooling vector will be then sent to the corresponding context-aware classifier to obtain the decision value.}
	\label{fig:attention}
\end{figure*}

\noindent\textbf{Context-aware attention}:
In the above attention generation function, the attention value is solely determined by $\mathbf{h}_{ij}$. Intuitively, however, it makes sense that different attention is required for different classification tasks. For example, to examine ``man riding bike'' and examine ``man playing football", different regions-of-interest should be focused on. We therefore propose to use a context-aware attention generator; i.e. we design $\mathbf{w}_{att}$ as a function of $(P, O1, O2)$. We can follow the framework in Eq.~(\ref{Eq:combine}) to calculate:
\begin{align}
	\mathbf{w}_{att}(P, O1,O2) = \mathbf{\bar{w}^a}_{p} + {{\mathbf{V}^a_p}}f(\mathbf{Q}E(O1,\,O2)),
\end{align}
where $\mathbf{\bar{w}^a}_{p}$ is the attention weight for the $p$-th interaction independent of its context and ${\mathbf{V}^a_p}$ transforms the semantic embedding of the context to the auxiliary attention weight for the $p$-th interaction. Note that in this case $\mathbf{w}_{att}$ depends on the interaction class $P$ and therefore different attention-pooling vectors $\widetilde {\mathbf{h}}_p$ will be generated for different $P$. $\widetilde {\mathbf{h}}_p$ will be then sent to the context-aware classifier for interaction $P$ to obtain the decision value for $P$ and the class that produces the maximal decision value will be considered as the recognized interaction. This structure is illustrated in Figure \ref{fig:attention}.

\subsection{Implementation details} \label{sec:implementation}
For all the above methods, we use the standard multi-class cross-entropy loss to train the models. The Adam algorithm~\cite{kingma2014adam} is applied as the optimization method. The methods that use appearance features involve convolutional layers from the standard VGG-16 network together with some newly added layers. For the former we initialize those layers with the parameters pretrained on ImageNet~\cite{russakovsky2015imagenet} and for the latter  we randomly initialize the parameters. We set the learning rate to 0.001 and 0.0001 for the new layers and VGG-16 layers respectively.

\section{Experiments}\label{sec:experiment}
To investigate the performance of the proposed methods, we analyse the effects of the context-aware interaction classifier, the attention models and various feature representations. Eight methods are implemented and compared:

\begin{enumerate}
	\item ``\textbf{Baseline1-app}'': We directly fine-tune the VGG-16 model to classify the interaction categories. Inputs are the union of subject and object boxes. This baseline models the interaction and its context separately, which corresponds to the approach described in Figure~\ref{fig:overview} (c).
	\item ``\textbf{Baseline1-spatial}'': We directly train a linear classifier to classify the spatial features described in Sec.~\ref{sec:spatial} into multiple interaction categories.
	\item ``\textbf{Baseline2-app}'':  We treat the combination of the interaction and its context as a single class and fine-tune the VGG-16 model for classification. This corresponds to using appearance feature to implement the method in Figure~\ref{fig:overview} (a).
	\item ``\textbf{Baseline2-spatial}'':  Similar to ``Baseline2-app''. We train a linear classifier to classify the spatial features into the classes derived from the combination of the interaction and its context.
	\item ``\textbf{AP+C}'': We apply the context-aware classifier to the appearance representation described in Sec.~\ref{sec:appearance}.
	\item ``\textbf{AP+C+AT}'': The basic attention-pooling representation described in Sec.~\ref{sec:attention} with the classifier in \textbf{AP+C}.
	\item  ``\textbf{AP+C+CAT}'': The context-aware attention-pooling representation described in Sec.~\ref{sec:attention} with the classifier in \textbf{AP+C}.
	\item ``\textbf{Spatial+C}'': We apply the context-aware classifier to the spatial features described in Sec.~\ref{sec:spatial}.

\end{enumerate}

Besides those methods, we also compare the performance of our methods against those reported in the related literature. However, it should noted that these methods may use different feature representation, detectors or pre-training strategies.

\subsection{Evaluation on the Visual Relationship dataset}

We first conduct experiments on the Visual Relationship Detection (VRD) dataset~\cite{lu2016visual}. This dataset is designed for evaluating the visual relationship ($\left\langle \emph{subject, predicate, object} \right\rangle$) detection, where the ``predicate'' in those datasets is equivalent to the ``interaction'' in our paper and we will use them interchangeably thereafter. It contains 4000 training and 1000 test images including 100 object classes and 70 predicates. In total, there are 37993 relationship instances with 6672 relationship types, out of which 1877 relationships occur only in the test set but not in the training set.

Following~\cite{lu2016visual}, we evaluate on three tasks: (1) For \textbf{predicate detection}, the input is an image and a set of ground-truth object bounding boxes. The task is to predict the possible interactions between pairs of objects. Since the interaction recognition is the main focus of this paper, the performance of this task provides the most relevant indication of the quality of the proposed method. (2) In \textbf{phrase detection}, we aim to predict $\left\langle \emph{subject-predicate-object} \right\rangle$ and localize the entire relationship in one bounding boxes. (3) For \textbf{relationship detection}, the task is to recognize $\left\langle \emph{subject-predicate-object} \right\rangle$ and localize both subject and object bounding boxes. Both boxes should have at least 0.5 overlap with the ground truth bounding boxes in order to be regarded as a correct prediction. For the second and third tasks, we use the object detection results (both bounding boxes and corresponding detection scores) provided in~\cite{lu2016visual}. This allows us to fairly compare the performance of the proposed interaction recognition framework without the influence of detection.

We use the Recall@100 and Recall@50 as our evaluation metric following~\cite{lu2016visual}. Recall@x computes the fraction of times the correct relationship is calculated in the top $x$ predictions, which are ranked by the product of the objectness confidence scores and the classification probabilities of the interactions. As discussed in~\cite{lu2016visual}, we do not use the mean average precision (mAP), which is a pessimistic evaluation metric because it cannot exhaustively annotate all possible relationships in an image.

\subsubsection{Detection results comparison}\label{sec:VRD}

In this section, we evaluate the performance of three detection tasks on the Visual Relationship Detection (VRD) benchmark dataset and provide the comprehensive analysis. We compare all the eight methods and the results in~\cite{sadeghi2011recognition, lu2016visual}. The results are shown in Table~\ref{tab:relationship}. From it we can make the following observations:

\noindent\emph{The effect of context-aware modeling:}
To validate the main point in this paper, we compare the proposed method against two context-interaction modeling baselines, i.e. baseline1-app, baseline2-app, baseline1-spatial and baseline2-spatial). By analysing the results, we can see that the proposed context-aware modeling methods (methods with ``AP'') achieves much better performance than the four baselines. The improvement achieved by use context-aware modeling is consistently observed for both spatial features and appearance features.  This justifies that the context information is crucial for interaction prediction.

\noindent\emph{Various feature representations:}
We also quantitatively investigate the performance of the proposed context-aware framework under various feature types. As can be seen in Table~\ref{tab:relationship}, the appearance feature representation performs consistently better than the spatial feature representation, especially for the baseline2 setting. This may be because the visual feature representation has richer discriminative power than the 14-dimensional spatial feature. Also, with our context-aware recognition framework, we can significantly boost the performance of both features and interestly in this case the gap between two types of features is largely diminished, e.g. AP+C+CAT vs. Spatial+C.

\noindent\emph{The effect of attention models:}
We also investigate the impacts of the attention scheme employed in our model by comparing AP+C, AP+C+AT and AP+C+CAT. The best results are obtained by utilizing the context-aware attention model. This justifies our postulate that it is better to make the network attend on the discriminative regions of feature maps.

\noindent\emph{Comparison with ~\cite{sadeghi2011recognition} and~\cite{lu2016visual}:}
Finally, we compare our methods with the methods in ~\cite{sadeghi2011recognition} and ~\cite{lu2016visual}. As seen, our methods achieve better performance than these two competing methods. Since our methods use the same object detection in~\cite{lu2016visual}, our result is most comparable to it. Note that our model does not employ explicit language priors modeling as in~\cite{lu2016visual} and our improvement purely comes from the visual cue. This again demonstrates the power of context-aware interaction recognition.

\begin{figure*}[!htb]
	\centering
	\resizebox{0.85\linewidth}{!}
	{\begin{tabular}{c}
			\includegraphics{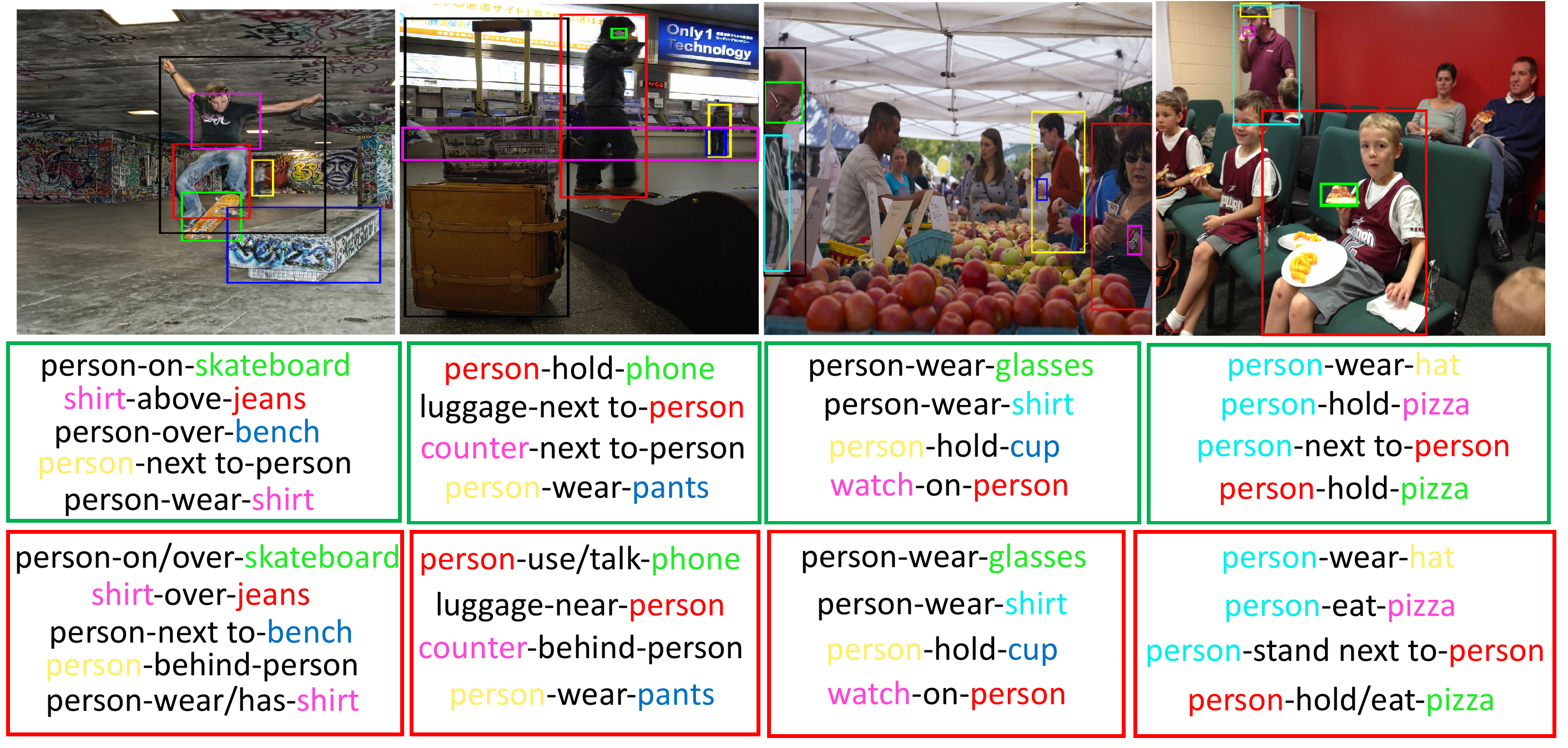}
		\end{tabular}
	}
	\caption{Qualitative examples of interaction recognition. We only predict the interaction between the ground-truth context bounding boxes. The phrases in the green bounding boxes are predicted while the phrases shown in the red bounding boxes are ground-truth.}
	\label{fig:qualitative1}
\end{figure*}

To better evaluate our approach, we further visualize some test examples of AP+C+CAT in Figure~\ref{fig:qualitative1}. We can see that our predictions are reasonable in most cases. %

\begin{table*}[t!]
	\centering
	\scalebox{0.8647}
	{
		\begin{tabular}{c|c c| c c c c}
			\hline
			\multirow{2}{*}{Method}&\multicolumn{2}{c|}{\textbf{Predicate Det.}} & \multicolumn{2}{c}{Phrase Det.} &\multicolumn{2}{c}{Relationship Det.}  \\  &R@100&R@50&R@100&R@50&R@100&R@50\\\hline
			Visual Phrase~\cite{sadeghi2011recognition} &1.91 &0.97 &0.07 &0.04 &- &-  \\
			Language Priors~\cite{lu2016visual} &47.87 &47.87  &17.03 &16.17 &14.70 &13.86 \\\hline
			Baseline1-app &18.13  &18.13  &6.02  &5.42  &5.54  &5.01   \\
			Baseline1-spatial &17.77 &17.77 &5.24 &4.77 &4.54 &4.19\\
			Baseline2-app &27.23 &27.23 &9.30 &7.91 &8.34 &7.03\\
			Baseline2-spatial &13.85 &13.85 &4.15 &3.06 &3.63 &2.63\\
			Spatial+C&51.17&51.17 &17.61 &15.46 &15.43 &13.51 \\
			AP+C&52.36 &52.36 &18.69 &16.91 &16.46 &14.88 \\
			AP+C+AT&53.12 &53.12 &19.08 &17.30 &16.89 &15.40 \\
			AP+C+CAT&\bf{53.59} &\bf{53.59} &\textbf{19.24}&\textbf{17.60} &\textbf{17.39} &\textbf{15.63}  \\\hline

		\end{tabular}}
		\caption{Evaluation of different methods on the visual relationship benchmark dataset. The results reported include visual phase detection (Phrase Det.), visual relationship detection (Relationship Det.)  and predicate detection (Predicate Det.) measured by Top-100 recall (R@100) and Top-50 recall (R@50). }
		\label{tab:relationship}
	\end{table*}

		\begin{table*}[!tbp]
			\centering
			\scalebox{0.8}
			{
				\begin{tabular}{c|c c c c c c c c}
					\hline
					\multirow{2}{*}{Method} & \multicolumn{2}{c}{Phrase Det.} &\multicolumn{2}{c}{Relationship Det.}&\multicolumn{2}{c}{Zero-Shot Phrase Det.}&\multicolumn{2}{c}{Zero-Shot Relationship Det.}\\  &R@100&R@50&R@100&R@50&R@100&R@50&R@100&R@50\\\hline
					
					CLC (CCA+Size+Position)~\cite{plummer2016phrase} &20.70 &16.89 &18.37 &15.08 &\textbf{15.23} &\textbf{10.86} &\textbf{13.43} &\textbf{9.67}\\
					VTransE~\cite{zhang2017visual} &22.42 &19.42 &15.20 &14.07 &3.51 &2.65 &2.14 &1.71\\
					Vip-CNN~\cite{li2017vip} &\textbf{27.91} &22.78 &20.01 &17.32 &- &- &- &-\\
					VRL~\cite{Liang2017VRD} &22.60 &21.37 &20.79 &18.19 &10.31 &9.17 &8.52 &7.94\\
					Faster-RCNN + (AP+C+CAT) &25.26 &23.88 &23.39 &20.14 &11.28 &10.73 &10.17 &9.57\\
					Faster-RCNN + (AP+C+CAT) + Language Priors &25.56 &\textbf{24.04} &\textbf{23.52} &\textbf{20.35} &11.30 &10.78 &10.26 &9.54\\\hline
					
				\end{tabular}}
				\caption{Results for visual relationship detection on the visual relationship benchmark dataset. Notice that we simply replace the detector with Faster-RCNN to extract a set of candidate object proposals without end-to-end jointly training the detector~\cite{zhang2017visual, li2017vip, Liang2017VRD} with the proposed method. And in CLC~\cite{plummer2016phrase}, they use features and detection results from Faster RCNN trained on external MSCOCO~\cite{lin2014microsoft} dataset and additional cues(e.g. size and position) are also incorporated.}
				\label{tab:VRD}
			\end{table*}

\subsubsection{ Zero-shot learning performance evaluation}\label{sec:zero-shot}
An important motivation of our method is to make the interaction classifier generalizable to unseen combinations of the interaction and context. In this section, we report the performance of our method on a zero-shot learning setting. Specifically, we train our models on the training set and evaluate their interaction classification performance on the 1877 unseen visual relationships in the test set. The results are reported in Table~\ref{tab:zeroshot}. From the table, we can see that the proposed methods work especially well in the zero-shot learning. For example, our best performed method (AP+C+CAT) almost doubled the performance on predicate detection in comparison with the Language Priors~\cite{lu2016visual} method. This big improvement can be largely attributed to the advantage of using the context-aware scheme to model the interaction. In the Language Priors~\cite{lu2016visual} method, the visual term for recognizing interaction is context-independent. Without context information to constrain the appearance variations,  the learned interaction classifier tends to overfit the training set and fails to generalize to images with unseen interaction-context combinations. In comparison, with context-aware modeling, we explicitly consider the visual appearance variations introduced by changing context, thus more accurate and generalizable interaction classifier can be learned.

One interesting observation made in Table~\ref{tab:zeroshot} is that the spatial feature representation produces better performance than the appearance based representation, as is evident from the superior performance of Spatial+C over AP methods. We speculate this is because spatial relationship features are more object independent and are less prone to overfiting the training set.

\begin{figure*}[!htb]
	\centering
	\resizebox{0.85\linewidth}{!}
	{\begin{tabular}{c}
			\includegraphics{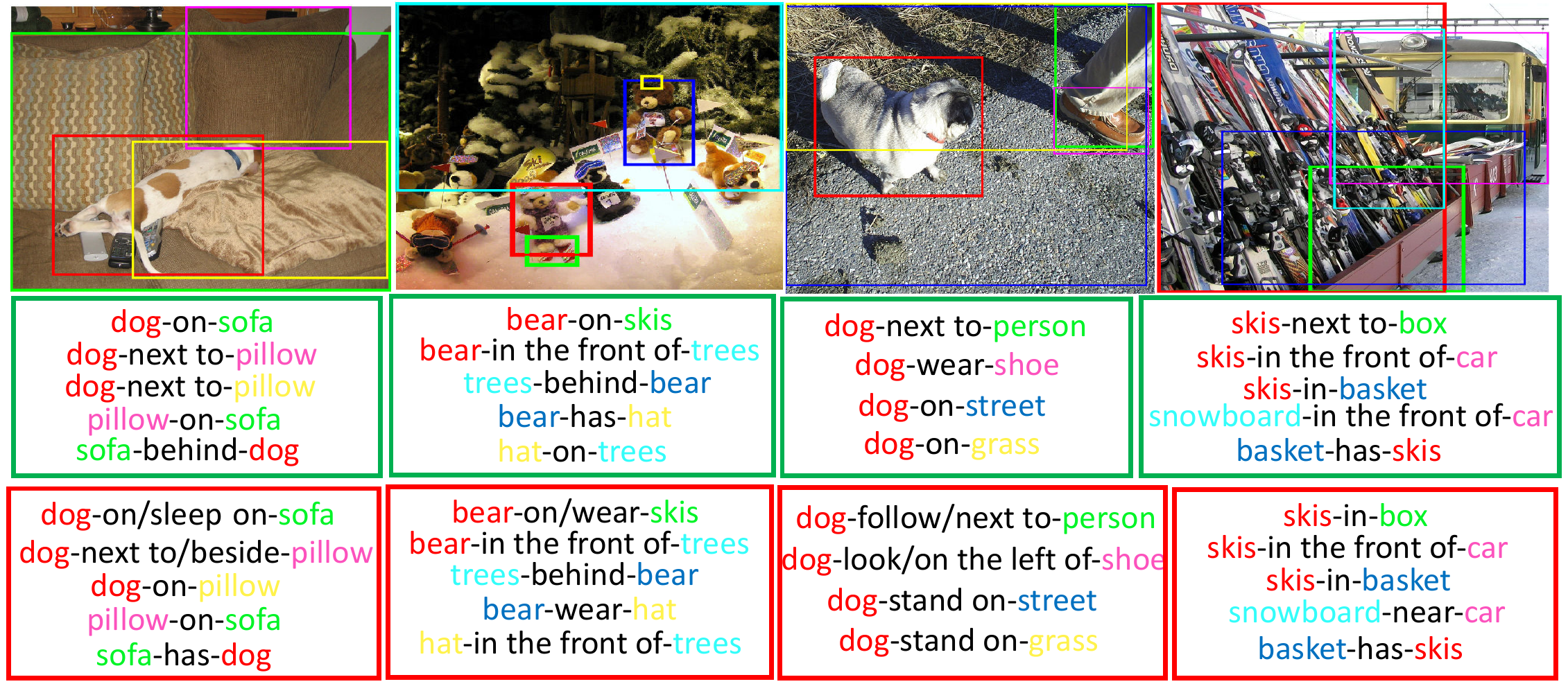}
		\end{tabular}
	}
	\caption{Qualitative examples of zero-shot interaction recognition. We only predict the interaction between the ground-truth context bounding boxes. The phrases in the green bounding boxes are predicted while the phrases shown in the red bounding boxes are ground-truth.}
	\label{fig:qualitative2}
\end{figure*}

To intuitively evaluate zero-shot performance, we add some test examples of AP+C+CAT in Figure~\ref{fig:qualitative2}. We can make reasonable predictions on unseen interaction-context combinations in most cases.

			\begin{table*}[h!]
				\centering
				\scalebox{0.87}
				{
					\begin{tabular}{c|c c |c c c c}
						\hline
						\multirow{2}{*}{Method}&  \multicolumn{2}{c|}{\textbf{Predicate Det.}} & \multicolumn{2}{c}{Phrase Det.} &\multicolumn{2}{c}{Relationship Det.} \\  &R@100&R@50&R@100&R@50&R@100&R@50\\\hline
						Language Priors~\cite{lu2016visual} &8.45 &8.45 &3.75 &3.36 &3.52 &3.13  \\\hline
						Baseline1-app&7.44 &7.44  &3.08  &2.82  &2.91  &2.74   \\
						Baseline1-spatial&7.27 &7.27 &2.14 &2.14 &2.14 &2.14\\
						Baseline2-app&7.36 &7.36 &2.22 &1.71 &2.05 &1.54\\
						Baseline2-spatial&0.43 &0.43 &0.09 &0.09 &0.09 &0.09\\
						Spatial+C &\bf{16.42} &\bf{16.42} &6.24 &5.82 &5.65 &5.30 \\
						AP+C&15.06 &15.06 &5.82 &5.05 &5.22 &4.62 \\
						AP+C+AT&15.00 &15.00 &5.62 &5.02 &5.36 &4.76 \\
						AP+C+CAT&16.37 &16.37&\textbf{6.59} &\textbf{5.99} &\textbf{5.99} &\bf{5.47}  \\\hline

					\end{tabular}}
					\caption{Results for zero-shot visual relationship detection on the visual relationship benchmark dataset. }
					\label{tab:zeroshot}
				\end{table*}

\subsubsection{Extensions and comparison with the state-of-the-art methods}

Since the main focus of above experiments is to validate the advantage of the proposed methods over four competing baselines, we did not explore some techniques which could potentially further improve the visual relationship detection performance on the VRD dataset. To make our method achieve more comparable performance on the visual relationship and visual phrase detection tasks, we may consider two straightfoward extensions for our method: (1) use a better detector and (2) incorporate the language term trained in~\cite{lu2016visual}. In the following part, we will examine the performance attained by applying these extensions and compare the resultant performance against the very latest state-of-the-art approaches~\cite{Liang2017VRD, li2017vip, zhang2017visual, plummer2016phrase} on the VRD dataset.

\noindent\emph{Improved detector:}
We first examine the effect of using a better detector by replacing the detection results obtained in~\cite{lu2016visual} with that obtained by a Faster-RCNN detector~\cite{girshick2015fast}. Note that the Faster-RCNN detector has also been used in~\cite{Liang2017VRD, li2017vip, zhang2017visual, plummer2016phrase} and using it will make our method comparable with the current state-of-the-arts. In our implementation, only the top 50 candidate object proposals, ranked by objectness confidence scores are extracted for mining relationships in per test image. The result of this modification is reported in Table~\ref{tab:VRD} with our method annotated as Faster-RCNN + (AP+C+CAT). As seen, our method achieves best performance on phrase detection R@50, relationship detection, zero-shot phrase and relationship detection. Note that our method can be further incorporated into the end-to-end relationship detection framework such as~\cite{li2017vip} to achieve even better performance.

\noindent\emph{Language priors:} Language priors make significant contribution to~\cite{lu2016visual} and in this section we apply the language priors released by~\cite{lu2016visual} to investigate its impact. Following~\cite{lu2016visual}, we multiply our best performed model Faster-RCNN + (AP+C+CAT) with the language priors for interactions to obtain the final detection scores and the result is shown in Table~\ref{tab:VRD} with the annotation Faster-RCNN + (AP+C+CAT) + Language Priors. Interestingly, the introduction of the language priors only introduces a marginal performance improvement. We suspect that is due to that our method builds a classifier with the information of both the interaction and context, and the correlation of interaction and context has been implicitly encoded. Therefore adding the language priors does not bring further benefit.

\subsection{Evaluation on the Visual Phrase dataset}
    \begin{table*}[h]
				\centering
				\scalebox{0.98}
				{
					\begin{tabular}{c|c c c c}
						\hline
						\multirow{2}{*}{Method} & \multicolumn{2}{c}{Phrase Detection} &\multicolumn{2}{c}{Zero-Shot Phrase Detection} \\  &R@100&R@50&R@100&R@50\\\hline
						Visual Phrase~\cite{sadeghi2011recognition} &52.7 &49.3 &- &- \\
						Language Priors~\cite{lu2016visual} &82.7 &78.1 &23.9 &11.4  \\\hline
						Baseline1-app&70.1 &65.6 &12.4 &10.5  \\
						Baseline1-spatial&68.3 &63.6 &10.3 &8.9\\
						Baseline2-app&77.5 &72.3 &11.0 &9.2\\
						Baseline2-spatial&15.7 &10.4 &1.1 &0.5\\
						Spatial+C&84.9 &80.8 &27.6 &15.7\\
						AP+C&85.9  &81.6  &28.5  &16.4  \\
						AP+C+AT &86.2  &82.1  &28.8  &17.9  \\
						AP+C+CAT&\textbf{86.8}  &\textbf{82.9} &\textbf{30.2}  &\textbf{18.7} \\\hline

					\end{tabular}}
					\caption{Comparison of performance on the Visual Phrase dataset.}
					\label{tab:VP}
				\end{table*}

Following~\cite{lu2016visual}, we also run additional experiments on the Visual Phrase~\cite{sadeghi2011recognition} dataset. It has 17 phrases, out of which 12 of these phrases can be represented as triplet relationships as in the VRD dataset. We use the setting of~\cite{lu2016visual} to conduct the experiment and report the R@50 and R@100 results in Table~\ref{tab:VP}. Since the Visual Phrase dataset does not provide detection results, we apply the RCNN~\cite{girshick2014rich} model to produce a set of candidate object regions and corresponding detection scores. As seen from Table~\ref{tab:VP}, AP+C+CAT again achieves the best performance. In comparison with the performance of~\cite{lu2016visual}, our method improves most in the zero-shot learning setting. This is consistent with the observation made in Sec.~\ref{sec:zero-shot}.

\section{Conclusion}
In this paper, we study the role of context in recognizing the object interaction pattern. After identifying the importance of using context information, we propose a context-aware interaction classification framework which is accurate, scalable and enjoys good generalization ability to recognize unseen context-interaction combinations. Further, we investigate various ways to derive the visual representation for interaction patterns and extend the context-aware framework to design a new attention-pooling layer. With extensive experiments, we validate the advantage of the proposed methods and produce the state-of-the-art performance on two visual relationship detection datasets.


\clearpage
\onecolumn

{
	\bibliographystyle{ieee}
	\bibliography{CSRef}
}

\end{document}